\documentclass[anonymous]{article}

\usepackage{spconf,amsmath,graphicx}

\usepackage{enumitem}
\setlist{nosep, leftmargin=14pt}

\usepackage{mwe} 

\usepackage{amsfonts}
\usepackage{bbold}
\usepackage{dsfont}
\usepackage{amsmath}
\usepackage{amssymb}
\usepackage{bbm}

\usepackage{CJKutf8}

\usepackage{cite}
\usepackage{amsmath,amssymb,amsfonts}
\usepackage{algorithmic}
\usepackage{graphicx}
\usepackage{textcomp}
\usepackage{hyperref}

\usepackage{makecell}
\usepackage{graphicx}
\usepackage{multirow}
\usepackage{mathrsfs}
\usepackage{bbding}

\usepackage{makecell}

\title{Unsupervised Domain Adaptation for Brain Vessel Segmentation through Transwarp Contrastive Learning}
\name{
    \begin{tabular}{c}
    Fengming Lin\textsuperscript{1 *},
    Yan Xia\textsuperscript{1 *}\thanks{\textasteriskcentered \quad Contribute equally to this work},
    Michael MacRaild\textsuperscript{1},
    Yash Deo\textsuperscript{1},
    Haoran Dou\textsuperscript{1},
    Qiongyao Liu\textsuperscript{1},
    Kun Wu\textsuperscript{1}, \\
    Nishant Ravikumar\textsuperscript{1 \dag},
    Alejandro F. Frangi\textsuperscript{1 2 \dag}\thanks{\dag \quad Joint last authors} \thanks{\dag \quad AFF is supported by the Royal Academy of Engineering INSILEX Chair (CiET1919/19), UKRI Frontier Research Guarantee INSILICO (EP/Y030494/1), and EC Sixth Framework Programme @neurIST (FP6-2004-IST-4-027703).}
    \end{tabular}%
}
\vspace{-0.5cm}
\address{
    \begin{tabular}{c}
    \textsuperscript{1} University of Leeds, \textsuperscript{2} University of Manchester  \quad
    \href{https://github.com/fmlinks/domain}{https://github.com/fmlinks/domain}
    \end{tabular}%
}

\begin{document}
%
\maketitle

\begin{abstract}
Unsupervised domain adaptation (UDA) aims to align the labelled source distribution with the unlabelled target distribution to obtain domain-invariant predictive models. Since cross-modality medical data exhibit significant intra and inter-domain shifts and most are unlabelled, UDA is more important while challenging in medical image analysis. This paper proposes a simple yet potent contrastive learning framework for UDA to narrow the inter-domain gap between labelled source and unlabelled target distribution. Our method is validated on cerebral vessel datasets. Experimental results show that our approach can learn latent features from labelled 3DRA modality data and improve vessel segmentation performance in unlabelled MRA modality data.

\end{abstract}


\section{Introduction}
Cerebral vessel segmentation offers invaluable assistance in precise diagnosis, surgical planning, monitoring disease progression, and evaluating treatment outcomes. In silico trials \cite{liu2023hemodynamics} using segmentation-derived vascular models enable the simulation of biological behaviours and support drug development efforts, ultimately enhancing the efficacy of personalised treatment strategies.
Traditional manual segmentation methods, while highly accurate, are labour-intensive and time-consuming. This has motivated the exploration of automated segmentation techniques \cite{lin2023high} as a more efficient alternative. However, the transition to computerised methods poses a new challenge: the reliance on fully labelled data for training.

\begin{figure}[!htb]
\centerline{\includegraphics[width=\columnwidth]{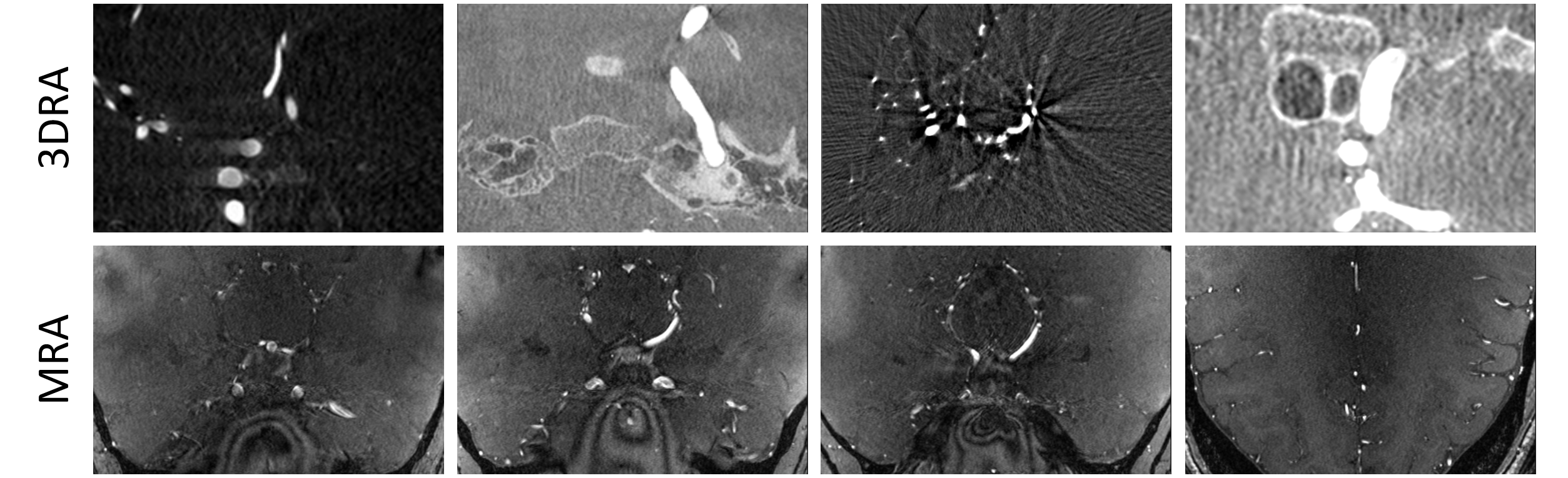}}
\caption{Visualization of 3DRA and MRA data reveals significant intra- and inter- domain shifts.}
\label{fig10}
\end{figure}
In scenarios where clinical vessel annotations in 3D Rotational Angiography (3DRA) \cite{bogunovic2011automated} modalities are ambiguous, semi-supervised learning methods  \cite{lin2023adaptive} with teacher-student structure \cite{yu2019uncertainty} leveraging both labelled and unlabeled patches have been proposed to address the issue. However, when the challenge escalates further, with the network being fed unlabelled Magnetic Resonance Angiography (MRA) \cite{deo2023learned, deo2023shape} data as input, semi-supervised methods become inadequate for supervision.
To address the challenge of data of different modes without labelled data, unsupervised domain adaptation (UDA) techniques, such as FDA \cite{yang2020fda}, DAFormer \cite{hoyer2022daformer}, HRDA \cite{hoyer2022hrda}, MIC \cite{hoyer2023mic}, and MSCDA \cite{kuang2023mscda}, have emerged. These methods transfer knowledge from well-annotated source domains to unlabelled target domains. 

These methods address the domain shift between source and target data.
However, the domain shift is considerable, even from inner source data in the context of cerebral vascular images. Fig.\ref{fig10} shows that 3DRA and MRA exhibit a substantial domain shift, with a significant domain shift even within the 3DRA modality when clinical data from different data centres are scanned. Mitigating domain shifts for segmentation across other modalities is more challenging due to the intricate nature of blood vessels, individual variabilities, and inherent noise and artefacts in imaging techniques.

Our work introduces a unique method integrating insights from unsupervised domain adaptation, semi-supervised learning, and contrastive learning to address significant domain shifts in cerebral vascular images. The key innovations are as follows:

\begin{itemize}
    \item[$\bullet$] This work presents an innovative symmetric adaptation network tailored for cross-modality segmentation of brain vasculature. This represents the first study to apply UDA techniques to the segmentation of the cerebral vessels from 3DRA to MRA.
    \item[$\bullet$] We introduce transwarp contrastive learning, a method that investigates features in the time and frequency domains within the latent space to achieve the alignment of content and style.
    \item[$\bullet$] Furthermore, we design a new homocentric squares Fourier domain adaptation to handle cross-domain nuisance variability without explicit feature alignment.
\end{itemize}

\section{Methods}
\begin{figure}[htb]
\centerline{\includegraphics[width=\columnwidth]{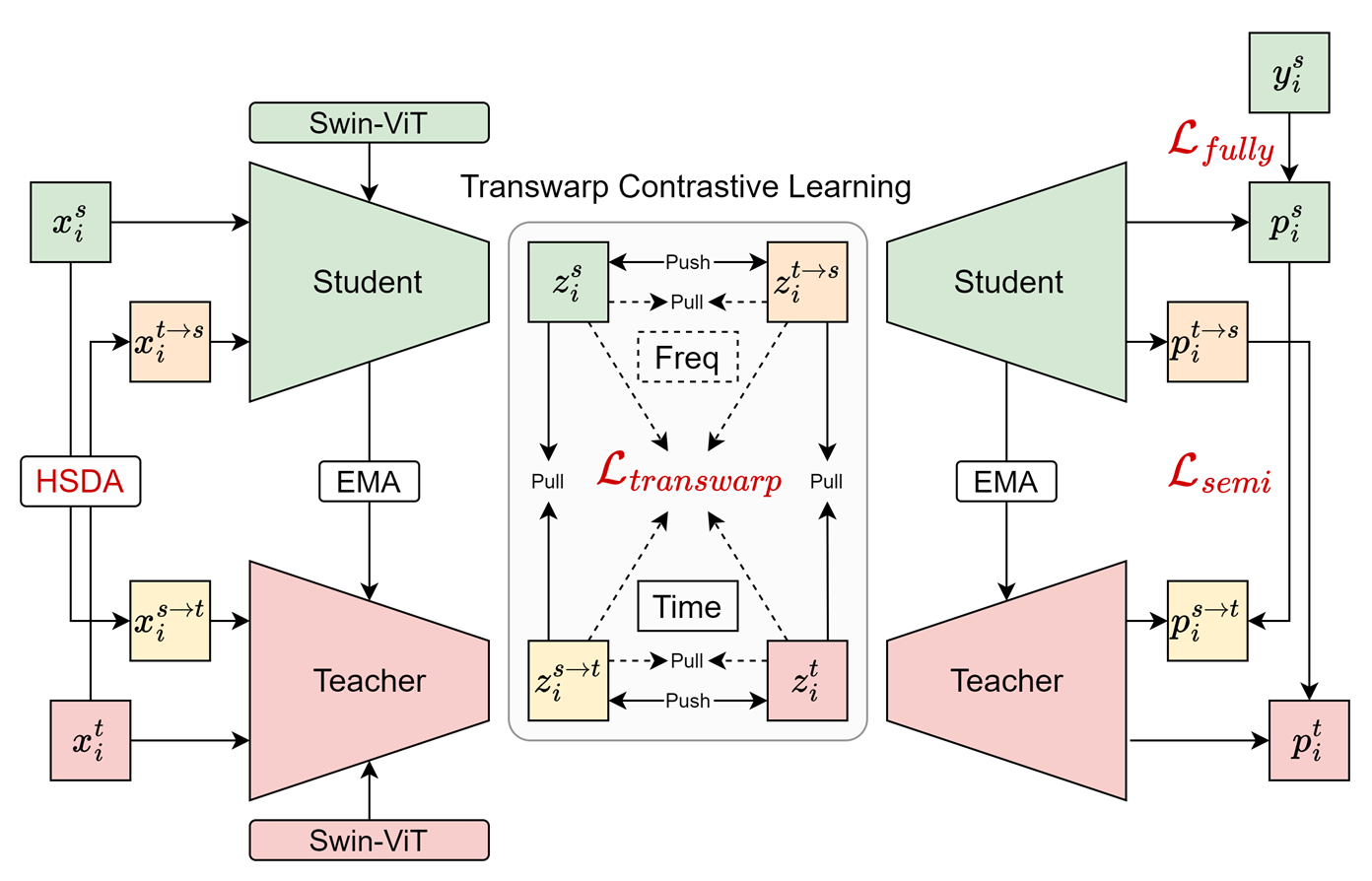}}
\caption{Schematic of the proposed method.
The method utilizes a composite loss function incorporating fully supervised, semi-supervised, and transwarp Contrastive Learning.
}
\label{fig1}
\end{figure}

This section presents our novel unsupervised domain adaptation approach (see Fig.\ref{fig1}) to learning instance-specific and domain-invariant representations. 

\subsection{Method Overview and Problem Formulation}

For image style transfer, FDA has demonstrated that low-frequency components of images represent style features  \cite{yang2020fda}. Therefore, we utilise the low-frequency components of latent features to extract style features from different inputs.
This novel paradigm in transwarp contrastive learning utilises a student-teacher network architecture, incorporating both content and style features using Fourier transform with a low-frequency mask, with the ultimate aim of narrowing the gap caused by different data modalities and magnifying the invariant feature extraction capabilities of the model.

As shown in Fig. \ref{fig1}, the student receives two inputs: labelled source domain data ($x_{i}^{s}$) and unlabelled target domain data undergone style transfer ($x_{i}^{t\rightarrow s}$). On the contrary, the teacher processes unlabelled target domain data ($x_{i}^{t}$) and style-transferred labelled source domain data ($x_{i}^{s\rightarrow t}$).
At the same time, we undertake the extraction of both content and style features. From the student and teacher networks, the characteristics of the content in the time domain are defined as $z_{i}^{s}$, $z_{i}^{t\rightarrow s}$, $z_{i}^{t}$ and $z_{i}^{s\rightarrow t}$. In parallel, we extract style features in the frequency domain, capturing the Fourier low-frequency attributes. These style features are expressed as $s_{i}^{s}$, $s_{i}^{t\rightarrow s}$, $s_{i}^{t}$, and $s_{i}^{s\rightarrow t}$.


\begin{figure}[!htb]
\centerline{\includegraphics[width=\columnwidth]{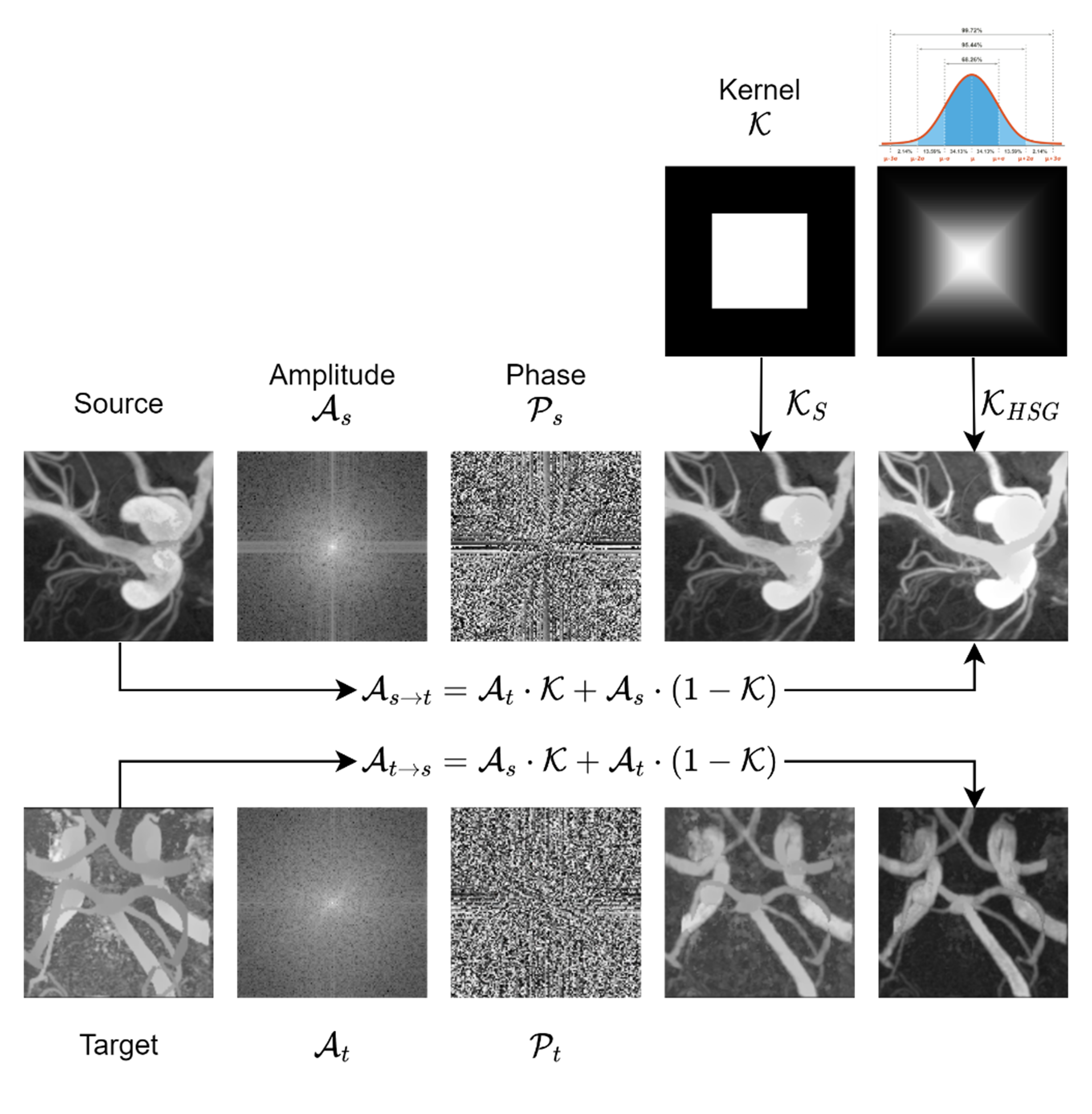}}
\caption{Homocentric squares Gaussian kernel $\mathcal{K}_{HSG}$ for image adaptation on 3DRA (source) and MRA (target) vessel patch.}
\label{hsda}
\end{figure}

The prediction of the student network on the source domain and the target on the source domain is denoted as $p_{i}^{s} $ and ${p}_{i}^{t\rightarrow s}$. We also denote the prediction of the teacher network in the target domain and the source in the target domain as $p_{i}^{t}$ and ${p}_{i}^{s\rightarrow t}$.
Our goal is to learn a task-specific network using a labelled source data set $\{ (x_{i}^{s}, y_{i}^{s})\}_{i=1}^{N_{s}}$ and an unlabelled target data set $\{(x_{i}^{t})\}_{i=1}^{N_{t}}$ to predict labels on test data from the target domain accurately.

\subsection{Homocentric Squares Domain Adaptation (HSDA)}


The purpose of pre-processing is to remove noise and facilitate the model extracting features specific to the vessels.
FDA proposes that the style of an image can be migrated by removing the low-frequency amplitude between the target image and the source image \cite{yang2020fda}. However, as observed in \cite{yang2020fda} \cite{basak2023semi}, the generated images consist of incoherent dark patches caused by abrupt changes in amplitude around the rectangular mask. 
Also, considering that the distribution of the spectrum is square instead of circles, we propose a 3D homocentric square Gaussian mask in Fig. \ref{hsda} to exchange the target and source amplitude to make a smooth transition in their style.

Let $\mathcal{F}_{A}(\cdot )$ and $\mathcal{F}_{P}(\cdot )$ be the amplitude and phase spectrum in frequency space of a 3D image, and $\mathcal{F}^{-1}$ indicate the inverse Fourier transform. We define a 3D homocentric square Gaussian mask $\mathcal{K}_{HSG}$ of the same size as $\mathcal{F}_{A}$, with $\sigma$ being the standard deviation. Given two patches $x_i^s$ and $x_i^t$, our proposed HSDA can be formulated as:
\begin{equation}
x_i^{s\rightarrow t} = \mathcal{F}^{-1}[\mathcal{F}_P(x_i^s), \mathcal{F}_A(x_i^t)\cdot \mathcal{K}_{HSG} + \mathcal{F}_A(x_i^s)\cdot (1- \mathcal{K}_{HSG})]
\end{equation}

\subsection{Fully-supervised Learning}

In the student network, labelled source samples $\left\{\left(x_{i}^{s}, y_{i}^{s}\right)\right\}_{i=1}^{N_{s}}$ are passed through the task-specific segmentation network to minimise the supervised loss $\mathcal{L}_{fully}$, which includes Dice similarity coefficient loss and cross-entropy:
\begin{equation}
\label{Lfully}
\begin{split}
\mathcal{L}_{fully}
&= \frac{1}{N_{s}} \sum_{i=1}^{N_{s}} \left ( 1 -  \frac{2\left | p_{i}^{s} \cap y_{i}^{s} \right | }{\left |  p_{i}^{s} \right | + \left | y_{i}^{s} \right |} -  y_{i}^{s}\log \left(p_{i}^{s}\right)  \right )
\end{split}
\end{equation}

\subsection{Semi-supervised Learning}
We employ semi-supervised learning, as the source data is labelled while the target data lacks labels. This involves using the predictions of the student network as pseudo-labels to supervise the output of the teacher network.

To perform alignment at the instance level, for the source patches, there are two predictions ${p}_{i}^{s}$ and ${p}_{i}^{s\rightarrow t}$. For the target patches, there are two predictions ${p}_{i}^{t}$ and ${p}_{i}^{t\rightarrow s}$.
First, for the student network, the model's ability to segment the source style is improved in fully supervised learning, where $p_i^s$ is supervised from $y_i^s$. After style migration, the student network also improves the segmentation performance of $p_i^{t\rightarrow s}$. We then use the output of the student network as a pseudo-label to supervise the output of the teacher network.
Specifically, we employ the mean square error defined in Eq. \ref{LMSE}  to reduce the discrepancy between the two predictions of the same patch.

\begin{equation}
\label{LMSE}
\mathcal{L}_{semi} = \frac{1}{N_{s}}\sum_{i=1}^{N_{s}}\left(p_{i}^{s}-{p}_{i}^{s\rightarrow t}\right)^2 + \frac{1}{N_{t}}\sum_{i=1}^{N_{t}}\left({p}_{i}^{t\rightarrow s}-p_{i}^{t}\right)^2
\end{equation}

\subsection{Transwarp contrastive learning}

Our objective in the time domain (content) is to enable the network to learn instance-specific features.
First, we aim to achieve positive content pairs (see Eq. \ref{posc}) by reducing the distance between content features derived from similar patches. The cosine similarity is computed to pull $z_{i}^{s}$ closer to $z_{i}^{s\rightarrow t}$ and $z_{i}^{t\rightarrow s}$ closer to $z_{i}^{t}$. As these pairs originate from identical patches and have only undergone processing through different networks, their latent features should closely align.
On the contrary, we set negative content pairs ( see Eq. \ref{negc}), derived from different modalities and patches, to maximise their inherent distance. This involves using the cosine similarity to push apart $c_{i}^{s}$ from $c_{i}^{t\rightarrow s}$ and similarly, distance $c_{i}^{s\rightarrow t}$ from $c_{i}^{t}$.

Venturing into the frequency domain (style), we aim for the network to become predominantly agnostic to modality regarding feature style. The fundamental objective is for the network to encapsulate the inherent vasculature data style, surpassing the intricacies of specific patches or modalities. The post-encoding outcome should display a more consistent style representation irrespective of the data's source, whether from various patches or modalities.

To achieve this goal, we take advantage of the low-frequency component of the latent feature as the style feature, aiming to ensure stylistic consistency across all segments. To accomplish this, we calculate the upbeat style pairs as outlined in Eq. \ref{poss}, aligning $s_{i}^{s}$, $s_{i}^{s\rightarrow t}$, $s_{i}^{t\rightarrow s}$, and $s_{i}^{t}$ to be near each other.
Furthermore, to preserve the integrity of the latent feature and mitigate excessive distortion, we restrict the number of encoder blocks to three, allowing for just two layers of max-pooling.

\begin{equation}
h(u, v)=\frac{u^{T} v}{\|u\|_{2}\|v\|_{2}}
\end{equation}

\begin{equation}
\label{posc}
pos_i^{c}=h\left(z_{i}^{s}, z_{i}^{s\rightarrow t}\right) + h\left(z_{i}^{t\rightarrow s}, z_{i}^{t}\right)
\end{equation}

\begin{equation}
\label{negc}
neg_i^{c}=h\left(z_{i}^{s}, z_{i}^{t\rightarrow s}\right)+h\left(z_{i}^{s\rightarrow t}, z_{i}^{t}\right)
\end{equation}

\begin{equation}
\label{poss}
\resizebox{\columnwidth}{!}{
$pos_i^{s} = h\left(s_{i}^{s}, s_{i}^{t\rightarrow s}\right) + h\left(s_{i}^{s}, s_{i}^{t}\right) + h\left(s_{i}^{s\rightarrow t}, s_{i}^{t}\right) + h\left(s_{i}^{s\rightarrow t}, s_{i}^{t\rightarrow s}\right)$
}
\end{equation}

\begin{equation}
\label{ltranswarp}
\mathcal{L}_{transwarp}=- \frac{1}{N_{}} \sum_{i=1}^{N_{}} \log \frac{(e^{pos_i^c} + e^{pos_i^s})/\tau}{e^{pos_i^c} + e^{pos_i^s} + e^{neg_i^c}}
\end{equation}

In summary, transwarp contrastive learning leverages Fourier transformation to extract temporal components as content features while utilising low-frequency components as style features. This approach enhances contrast in content aspects while reducing contrast in style aspects and enhancing network adaptation capabilities across domains.

\subsection{Overall framework and training objective}
The weight ratio of fully supervised loss is higher than that of semi-supervised loss and contrastive loss of the transwarp (8:1:1).
We prioritise the fully supervised loss to ensure training robustness and prevent the network from becoming overly confident and introducing noise during the initialisation stage.

\begin{equation}
\mathcal{L}=\lambda_1 \cdot \mathcal{L}_{fully} + \lambda_2 \cdot \mathcal{L}_{semi} + \lambda_3 \cdot \mathcal{L}_{transwarp}
\end{equation}

\section{Experiments}

\subsection{Datasets}

Our study trained the network using labelled 3DRA and unlabelled MRA from the Aneurist data set and tested it with the SMILE data set.
\textbf{Aneurist-3DRA} \cite{benkner2010neurist}:
This dataset includes 223 partially annotated 3D brain vessel images from four different centres, each using unique scanners and imaging protocols. 
\textbf{Aneurist-MRA} \cite{benkner2010neurist}:
Consisting of 207 high-quality clinical MRA data, this dataset lacks vascular segmentation labels and is not paired with Aneurist-3DRA.
\textbf{SMILE-UHURA} \cite{SMILEUHURA}:
This dataset offers 14 fully labelled cerebral vessel cases in MRA. ToF angiography images were initially pre-segmented automatically and underwent extensive manual refinement to ensure accuracy.

\subsection{Experimental Setup}


Our experiments used an NVIDIA GeForce RTX 3090 GPU (24GB VRAM) with 128GB RAM. We use Swin-UNet \cite{cao2022swin} for both student and teacher networks, using patch-based learning with patch size [128, 128, 128]. We aim to have 100 training epochs with optimisation using the Adam optimiser. Data augmentation techniques, including random rotational adjustments and flipping, were applied. 
The initial learning rate is 0.001 and decreases by a factor of 0.1 every ten epochs.
Teacher network parameters $\theta_{tea}$ received standard updates, while student network parameters $\theta_{stu}$ were updated using the exponential moving average approach in Eq. \ref{eq13}. 

\begin{equation}
\theta_{{stu}} = \alpha \cdot \theta_{{stu}} + (1 - \alpha) \cdot \theta_{{tea}}
\label{eq13}
\end{equation}

\begin{equation}
\alpha = \min \left(1 - \frac{1}{{iter + 1}}, decay\right)
\label{eq14}
\end{equation}

\subsection{Quantitative Results}

In Table \ref{table1}, we present a summary of segmentation performance compared to state-of-the-art (SOTA) techniques, which encompasses various scenarios: (1) Source-only learning on the target image, (2) Different UDA methods, and (3) Fully supervised learning with labelled target data. 
MSCDA is designed for medical images in UDA methods, and DAFormer, MIC, and HRDA are natural image UDA methods. Our approach excels in critical metrics, including the Dice similarity coefficient (DSC), Sensitivity (Sen), Jaccard index (Jac) and Volume similarity (VS), highlighting its effectiveness in detecting positive instances, achieving overall agreement, and ensuring volume matching accuracy. While we outperform other domain adaptation methods and the source-only learning baseline, there is room for further enhancement in our unsupervised approach compared to fully supervised methods.

\begin{figure}[!ht]
\centerline{\includegraphics[width=\columnwidth]{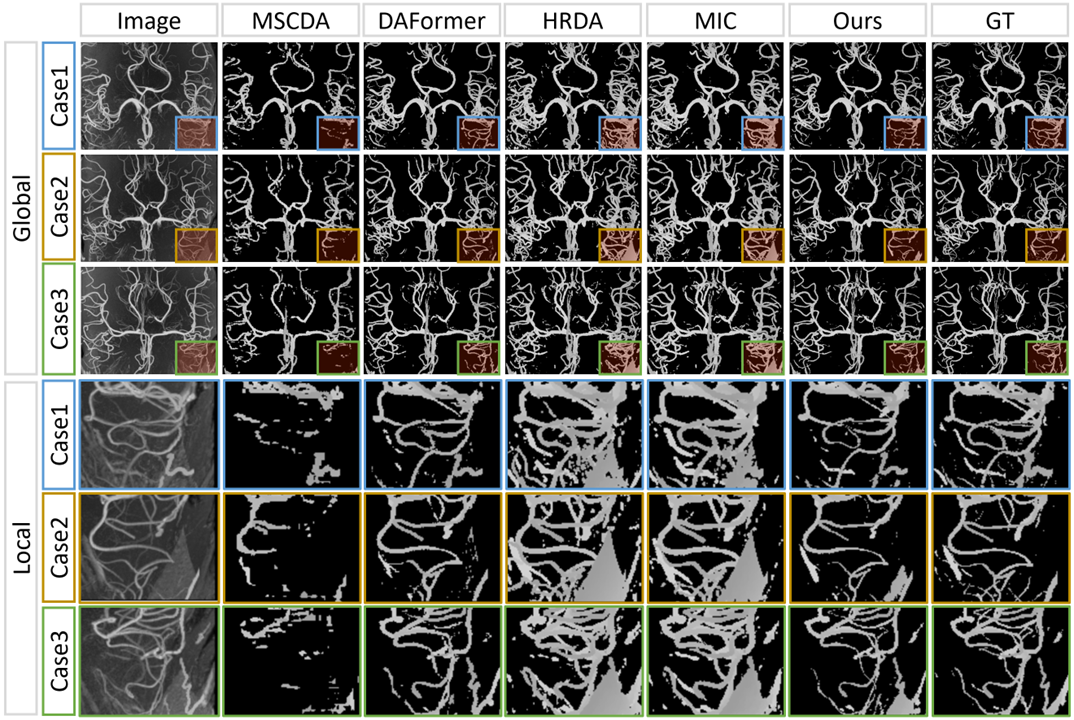}}
\caption{Visualisation comparison on MIP maps. Ours shows less over-segmentation on local area.}
\label{fig05}
\end{figure}

\begin{table}[!ht]
\centering
\caption{Comparison of Segmentation Performance with UDA SOTAs and different training strategy.* indicates $p < 0.05$ in t-test.}
\label{table1}
\resizebox{\columnwidth}{!}{
\begin{tabular}{c|c|c|c|c}
\hline
\hline
Methods&DSC (\%) $\uparrow$
&Sen (\%) $\uparrow$
&Jac (\%) $\uparrow$
&VS (\%) $\uparrow$
\\
\hline
$\mathcal{S} \rightarrow \mathcal{T}$\cite{lin2023high}
& \makecell[l]{31.48 ± 6.76}
& \makecell[l]{18.89 ± 5.00}
& \makecell[l]{18.88 ± 5.00}
& \makecell[l]{31.52 ± 6.75}
\\
\hline
MSCDA\cite{kuang2023mscda}
& \makecell[l]{41.18 ± 4.70}
& \makecell[l]{27.57 ± 4.96}
& \makecell[l]{26.04 ± 3.84}
& \makecell[l]{49.12 ± 8.69}
\\
DAFormer\cite{hoyer2022daformer} 
& \makecell[l]{57.75 ± 6.35}
& \makecell[l]{42.84 ± 8.07}
& \makecell[l]{40.89 ± 6.52}
& \makecell[l]{63.37 ± 9.70}
\\
MIC\cite{hoyer2023mic}
& \makecell[l]{67.16 ± 2.02}
& \makecell[l]{59.07 ± 7.16}
& \makecell[l]{50.59 ± 2.27}
& \makecell[l]{84.18 ± 9.49}
\\
HRDA\cite{hoyer2022hrda}
& \makecell[l]{68.35 ± 2.74}
& \makecell[l]{60.03 ± 8.57}
& \makecell[l]{51.98 ± 3.14}
& \makecell[l]{83.31 ± 9.68}
\\
Ours
& \makecell[l]{\textbf{72.65 ±} \textbf{6.65} *}
& \makecell[l]{\textbf{64.75 ±} \textbf{8.06} *}
& \makecell[l]{\textbf{57.46 ±} \textbf{7.80} *}
& \makecell[l]{\textbf{85.47 ±} \textbf{9.65} *}
\\
\hline
$\mathcal{T} \rightarrow \mathcal{T}$\cite{cao2022swin}
& \makecell[l]{79.76 ± 1.92}
& \makecell[l]{74.61 ± 7.77}
& \makecell[l]{66.37 ± 2.69}
& \makecell[l]{90.06 ± 5.74}
\\
\hline
\hline
\end{tabular}
}
\end{table}

Table \ref{table2} presents an ablation study that builds upon the foundation of the FDA and progressively integrates various components into our approach. As we observe improvements in the experimental results, it becomes evident that each component in our method contributes positively to the outcome.

\begin{table}[!htbp]
\centering
\caption{Ablation Study: Gradual Addition of Components from Top to Bottom. 
}
\label{table2}
\resizebox{\columnwidth}{!}{
\begin{tabular}{c|c|c|c|c}
\hline
Components &DSC (\%) $\uparrow$&Sen (\%) $\uparrow$&Jac (\%) $\uparrow$&VS (\%) $\uparrow$
\\
\hline
$\mathcal{L}_{fully}$
&\makecell[l]{61.84 ± 7.08}
&\makecell[l]{46.29 ± 8.48}
&\makecell[l]{45.16 ± 7.77}
&\makecell[l]{64.88 ± 8.47}
\\
$\mathcal{L}_{semi}$
&\makecell[l]{64.60 ± 7.36}
&\makecell[l]{49.08 ± 8.75}
&\makecell[l]{48.00 ± 8.17}
&\makecell[l]{67.48 ± 8.42}
\\
$\mathcal{L}_{transwarp}$
& \makecell[l]{{67.55} ± {6.81}}
& \makecell[l]{{52.75} ± {8.65}}
& \makecell[l]{{50.95} ± {7.67}}
& \makecell[l]{{72.16} ± {8.47}}
\\
Ours HSDA
& \makecell[l]{{72.65 ±} {6.65}}
& \makecell[l]{{64.75 ±} {8.06}}
& \makecell[l]{{57.46 ±} {7.80}}
& \makecell[l]{{85.47 ±} {9.65}}
\\
\hline
\end{tabular}}
\end{table}
\vspace{-0.3cm}

\subsection{Visual Inspection}
In addition to conducting quantitative comparisons, we also performed visual comparisons of results with similar numerical values in Table \ref{table1}.
In Fig. \ref{fig05}, we visually compared maximum intensity projection (MIP) maps between our method and the top-performing algorithm MIC and HRDA from UDA-SOTAs. It is evident from the image that within the red bounding box, MIC and HRDA tends to exhibit over-segmentation. Its capacity to distinguish between vessels and brain tissue noise is significantly inferior to our approach.

\section{Conclusion}

In conclusion, our novel transwarp contrastive learning framework represents a significant advancement in the field of UDA in the segmentation of brain vessels. We have effectively managed the challenges posed by diverse data modalities by utilising content pairs from the time domain and style pairs from the frequency domain. 
This innovative approach not only facilitates knowledge transfer from source domains with domain shift but also ensures exceptional precision in feature extraction for application in target domains. Future work will focus on enhancing domain generalisation capabilities.

\bibliographystyle{IEEEbib}
\bibliography{refs}

\end{document}